\pdfoutput=1

\documentclass[11pt]{article}

\usepackage{acl}

\usepackage{times}
\usepackage{latexsym}
\usepackage{multirow}
\usepackage{graphicx}
\usepackage{caption}
\usepackage{subcaption}
\usepackage{amsmath}
\usepackage{multirow}
\usepackage{pifont}
\newcommand{\cmark}{\ding{51}}%
\newcommand{\xmark}{\ding{55}}%

\usepackage[T1]{fontenc}

\usepackage[utf8]{inputenc}

\usepackage{microtype}

\title{Finding the Right Recipe for Low Resource Domain Adaptation in Neural Machine Translation}

\author{Virginia Adams, Sandeep Subramanian, Mike Chrzanowski, \\ {\bf Oleksii Hrinchuk}, {\bf Oleksii Kuchaiev} \\
         NVIDIA \\ Santa Clara, CA \\ 
         \textit{\{vadams, sandeepsub, mchrzanowski, ohrinchuk, okuchaiev\}@nvidia.com}} 
\begin{document}
\maketitle
\begin{abstract}
General translation models often still struggle to generate accurate translations in specialized domains. To guide machine translation practitioners and characterize the effectiveness of domain adaptation methods under different data availability scenarios, we conduct an in-depth empirical exploration of monolingual and parallel data approaches to domain adaptation of pre-trained, third-party, NMT models in settings where  architecture change is impractical. We compare data centric adaptation methods in isolation and combination. We study method effectiveness in very low resource (8k parallel examples) and moderately low resource (46k parallel examples) conditions and propose an ensemble approach to alleviate reductions in original domain translation quality. Our work includes three domains: consumer electronic, clinical, and biomedical and spans four language pairs - Zh-En, Ja-En, Es-En, and Ru-En. We also make concrete recommendations for achieving high in-domain performance and release our consumer electronic and medical domain datasets for all languages and make our code publicly available.  
\end{abstract}

\section{Introduction}
The prevalence of pre-trained models has fueled exciting academic and industry progress in natural language processing. It has allowed practitioners to re-use computationally expensive training steps and bypass the most inaccessible portion of model training \cite{wolf2019huggingface}. In neural machine translation (NMT), these general pre-trained models still struggle with translating domain specific material and require further tuning to achieve desired performance. In this work, we focus on methods for adapting off-the-shelf, third party, pre-trained translation models in which no additional architectural changes or edits to the model's pre-training scheme are possible. Intuitively, domain adaptation using clean, in-domain parallel data should provide the best results. However, such data is often hard and expensive to obtain. Monolingual in-domain data is more abundant and, at the cost of translation quality, can be used to generate synthetic parallel data.

We aim to elucidate which domain adaptation approaches of off-the-shelf translation models best suit various low data resource scenarios to yield the highest in-domain translation quality. We explore the benefits and trade-offs of domain adaptation methods in combination and isolation. While setting up our experiments, we found English in-domain monolingual data to be much more readily available than in-domain data for other languages. Collecting high quality monolingual electronic, medical, and biomedical domain data for adapting out of English translation models (En$\rightarrow$*) proved to be difficult to the extent that we limit our study to models translating into English (*$\rightarrow$En). For all experiments, the source language is one of Russian, Chinese, Spanish, or Japanese and the target language is always English. Similarly, as the source language is always non-English, we limit the scope of our work to scenarios with differing access to in-domain parallel and target side monolingual data. 

\begin{table*}
\centering
\begin{tabular}{|c|c|c|c|c|c|c|c|c|c|}
\hline
\multirow{2}{*}{\textbf{This Study}} & \multicolumn{3}{c|}{\textbf{In-Domain Data Scenario}} & \multicolumn{6}{c|}{\textbf{Adaptation Approaches}} \\
\cline{2-10}
& Parallel & Source Mono & Target Mono & FT & SF & BT & ST & TBT & TST \\
\hline
\cmark & \cmark & \xmark & \xmark & \cmark & \cmark & \xmark & \xmark & \xmark & \xmark \\
\hline
\xmark  & \cmark & \cmark & \xmark & \cmark & \cmark & \xmark & \cmark & \xmark & \cmark \\
\hline
\cmark & \cmark & \xmark & \cmark & \cmark & \cmark & \cmark & \xmark & \cmark & \xmark \\
\hline
\xmark & \cmark & \cmark & \cmark & \cmark & \cmark & \cmark & \cmark & \cmark & \cmark \\
\hline
\xmark & \xmark & \cmark & \cmark & \xmark & \cmark & \cmark & \cmark & \xmark & \xmark \\
\hline
\xmark & \xmark & \cmark & \xmark & \xmark & \xmark & \xmark & \cmark & \xmark & \xmark \\
\hline
\cmark & \xmark & \xmark & \cmark & \xmark & \cmark & \cmark & \xmark & \xmark & \xmark \\
\hline
\end{tabular}
\caption{\label{scenarios} Data Resource Scenarios and Corresponding Possible Adaptation Methods. Adaptation approaches include 1) FT - Finetuning, 2) SF - Shallow Fusion decoding with in-domain language models, 3) BT - Backtranslation, 4) ST - Self-training, 5) TBT - Tagged Backtranslation, 6) TST - Tagged Self-training}
\end{table*}

We examine domain adaptation approaches under three in-domain data availability scenarios: parallel data only, target side monolingual data only, and both parallel and target side monolingual data. We compare parallel in-domain fine-tuning, mixed-domain fine-tuning \citep{zhang-etal-2019-curriculum}, traditional back-translation \citep{sennrich-etal-2016-improving, edunov2018understanding}, tagged back-translation \citep{caswell-etal-2019-tagged}, and in-domain language model shallow fusion across scenarios where applicable. See Table \ref{scenarios} for a breakdown of data availability conditions and the fixed architecture adaptation methods that can be applied to each. 

Further, we use of domain classifiers to mine additional in-domain parallel data - adding dimension to the quantity verses quality trade off encountered in back-translation discussions. Finally, we suggest an ensemble approach to mitigate degradation in original domain performance.

\section{Contributions}
Our main contributions include:
\begin{itemize}
    \item A systematic empirical comparison of domain adaptation approaches of third-party, fixed architecture transformer-based NMT models
    \item A simple ensemble method to preserve original domain performance while gaining translation ability across new domains 
    \item An effective low resource parallel data augmentation approach to improve in-domain performance
    \item The release of consumer electronic and clinical domain datasets across Russian $\rightarrow$ English, Chinese $\rightarrow$ English, Spanish $\rightarrow$ English, and Japanese $\rightarrow$ English translation pairs and our code.
\end{itemize}

\section{Related Work}
\subsection{Low-Resource Machine Translation}
Solutions to low resource machine translation range from transfer learning approaches and data mining strategies like the ones we focus on in this paper, to meta learning approaches \citet{gu-etal-2018-meta}, pre-training methods \citet{song2019mass}, multilingual \citet{zoph-etal-2016-transfer} strategies, and methods making strategic use of monolingual source and target data \cite{zhang-zong-2016-exploiting}. \citet{NIPS2016_5b69b9cb} proposed a dual learning reinforcement learning solution to low-resource machine translation that takes advantage of large amounts of both source and target monolingual data. \citet{AhmadniaDorr+2019+268+278} combines dual learning with self-training and co-learning using the synthetic translation examples produced during the dual learning round trip translations to further train their translation models. 
 
\subsection{Domain Adaptation Strategies}
Domain adaptation is a widely studied research area with strategies that vary by data access, training objective, and architectural changes. As we focus purely on data centric strategies for adapting off-the-shelf NMT models, so other adaptation approaches like adapter methods \citet{houlsby2019parameter}, differential adaptation, and deep fusion\citet{dou2019domain} that require access and/or edits to the NMT model's architecture fall out of the scope of this work. In section \ref{our-domain-adapt-methods} we give a detailed introduction to each domain adaption method we explore. 

\subsection{Empirical Studies of Fixed-Architecture Domain Adaptation}
There are a couple of existing empirical comparisons of domain adaptation methods using LSTM neural machine translation models. \citet{chu-etal-2017-empirical} explores mixed domain fine-tuning and compares different in-domain up-sampling strategies to mitigate overfitting on generally low resource parallel domain data. Our work is most similar to that of \citet{Chenhui-Chu2018}. In their empirical study, \citet{Chenhui-Chu2018} compares fine-tuning NMT models on parallel mixed domain data with fine-tuning models on data that was synthetically generated via back-translation. Though they propose a single domain adaptation method for RNN based models in which they combine back-translation, mixed-domain fine-tuning, and shallow fusion strategies, they do not explore iterative combinations of these approaches and therefore do not give strong evidence for one method over another. They also don't consider tagged back-translation, multi-domain ensembling, or additional data mining strategies as we do in this work.

\citep{saunders2021domain} and \cite{chu-wang-2018-survey} perform literary surveys on domain adaptation approaches for neural machine translation. Other works have explored domain adaptation under one of the three situations we compare in our investigation. \citet{Sun2019AnES} studies training and adapting unsupervised translation models with exclusively monolingual data. They use cross-lingual language model pre-training \cite{NEURIPS2019_c04c19c2} to initialize their unsupervised neural machine translation (UNMT) models, then train and fine-tune their models according to different scenarios modulating the presence or absence of in-domain and out-of-domain source and target monolingual data. 

\section{Methods} \label{our-domain-adapt-methods}
We focus on the efficacy of domain adaptation approaches for pre-trained models with access to different combinations of parallel and monolingual target language data. We assume access to out-of-domain NMT models in both language directions, but narrow our study to improving in-domain performance in the Other Language $\rightarrow$ English direction, using English $\rightarrow$ Other Language models solely for back-translation. We empirically compare domain adaptation methods separately and together. We only consider adaptation of a fixed-architecture base models. 

\subsection{Fine-Tuning}
We characterize the compromise between minimizing general domain degradation and improving in-domain performance in our parallel data approaches. We also experiment with fine-tuning baseline models on solely parallel in-domain data and on a mix of original and in-domain data \citep{zhang-etal-2019-curriculum}.

\subsection{Back-Translation}
In back-translation \citep{sennrich-etal-2016-improving, edunov2018understanding, lample2018unsupervised}, target side monolingual data is used to generate synthetic parallel data. A reverse direction translation model translates the target language into the source language, often using sampling instead of greedy decoding to increase translation diversity. The forward direction translation model is fine-tuned on this generated parallel data. The reverse direction translation model can be used as is, or fine-tuned with available domain data before back-translation \citep{kumari-etal-2021-domain, artetxe-etal-2018-unsupervised}. In tagged back-translation \citep{caswell-etal-2019-tagged} a special token (e.g. <BT>) is prepended before the synthetically generated source sentence. This tag helps the model differentiates noisy synthetic translations from ground truth examples.

\subsection{Shallow Fusion Decoding}
Shallow fusion \citep{gulcehre2015using, lample2018unsupervised, DBLP:journals/corr/abs-1910-02555} combines the next token probability predicted by a pre-trained language model possessing parameters $\phi_{t}$ with the next token probability predicted by the NMT model's decoder $\theta_{t}$ at every time step $t$. The generated translation benefits from the fluidity and target language knowledge of the language model while relying on the NMT decoder for semantic content. The two probabilities are added with a language model coefficient $\lambda_{LM}$ scaling the language model's contribution.

\begin{equation}
\begin{split}
    P(y_t|y_{<t},x) = P_{NMT}(y_t|y_{<t},x;\theta_{t}) \\
                      + \lambda_{LM} * P_{LM}(y_t|y_{<t};\phi_{t})
\end{split}
\end{equation}

The language model is fine-tuned on target side monolingual data before shallow fusion decoding. 

\subsection{Ensemble}
We propose using an ensemble of fine-tuned models with the base translation model to gain the benefits of adaptation across domains while maintaining high original domain performance. $k$ indicates the total number of models in the ensemble, we average their probability distributions over the next token at every decoding time step $t$.  

$$P(y_t|y_{<t},x;\theta_{1} \ldots \theta_{k}) = \frac{1}{k} \sum_{i=1}^k P(y_t|y_{<t},x;\theta_{i})$$

Here $P(y_t|y_{<t},x;\theta_{i})$ is the probability of target token $y$ at time step $t$ for a single NMT model $i$ given the input tokens $x$ and previously generated tokens $y_{<t}$. 

\section{Datasets}
We define low resources as falling between 5k and 9k in-domain parallel training examples and moderately low resource has an order of magnitude more data, in this case around ~47k examples. We create low resource consumer electronic and medical domain datasets for each language pair. We also gathered in-domain monolingual data for the medical and consumer electronic domains. Final data totals for each language, split, and domain are listed in Table \ref{data-totals}. We make the datasets and dataset creation code publicly available. \footnote{Anonymized} 

\begin{table}
\centering
\resizebox{\columnwidth}{!}{
\begin{tabular}{|c|c|c|c|c|}
\hline
\textbf{Domain} & \textbf{Language Pair} & \textbf{Train} & \textbf{Val} & \textbf{Test} \\
\hline
\multirow{ 4}{*}{Electronic} & Zh $\rightarrow$ En & 7,041 & 475 & 479 \\
& Ja $\rightarrow$ En & 6,777 & 452 & 460 \\
& Es $\rightarrow$ En & 6,973 & 421 & 430 \\
& Ru $\rightarrow$ En & 7,276 & 478 & 522 \\
\hline
\multirow{ 4}{*}{Medical} & Zh $\rightarrow$ En & 8,760 & 448 & 446 \\
& Ja $\rightarrow$ En & 5,399 & 460 & 461 \\
& Es $\rightarrow$ En & 8,494 & 434 & 437 \\
& Ru $\rightarrow$ En & 5,401 & 507 & 493 \\
\hline
\multirow{ 1}{*}{Biomedical} & Ru $\rightarrow$ En & 46,782 & 279 & - \\
\hline
\end{tabular}}
\caption{\label{data-totals}
Total parallel examples for each split of each language pair.
}
\end{table}
\subsection{Parallel Consumer Electronic Dataset}

We used human generated translations from consumer electronic websites to construct the consumer electronic dataset. We crawled multilingual versions of XXXX\footnote{Website anonymized for review} website, matching translated versions of each page via their URLs. 

To convert document level translations into aligned sentences, we separated sentences using NLTK's sentence splitter \footnote{https://www.nltk.org/api/nltk.tokenize.html} for English, Spanish, and Russian. We used the Spacy \footnote{https://spacy.io/models/zh} library's Chinese splitter to separate Mandarin sentences and the Konoha \footnote{https://github.com/himkt/konoha} library to split Japanese sentences. We then used the Vecalign library \footnote{https://github.com/thompsonb/vecalign} \citep{thompson-koehn-2019-vecalign} in conjunction with the Language-Agnostic SEntence Representations (LASER) multilingual embedding library \citep{artetxe-schwenk-2019-massively} to align translated document pairs on a sentence level. We selected sentence pairs within a set cosine distance range of 0.07 to 0.6 for the training split, where we define cosine distance as (1 - cosine similarity). For the validation and test splits, we used a narrower cosine distance range of 0.1 to 0.5 and removed overlapping validation and test examples from the train split. We manually cleaned the validation and test splits-- separating examples containing multiple sentences and removing sentence fragments lacking a clear meaning.  

\subsection{Parallel Medical Dataset}
Parallel translations of medical domain data were gathered from translated pdfs publicly provided by the NIH U.S. National Library of Medicine \footnote{https://medlineplus.gov/languages/languages.html}. An identical process to the one used for the consumer electronic dataset was employed to create the parallel medical train, validation, and test splits.

\subsection{Parallel Biomedical Dataset}
We use the publicly available WMT'20 biomedical shared task train split for our Ru $\leftrightarrow$ En biomedical domain data. To explore the benefits of noisy parallel data, we also mine additional parallel in-domain data from the out-of-domain En $\leftrightarrow$ Ru WMT'21 News dataset. Here, noise comes from potential domain misclassification instead of from erroneous translation as with back-translation.

To collect this data, we trained English and Russian biomedical domain classifiers. Each classifier utilized a pre-trained BERT Base style encoder \cite{devlin2018bert} with added classification layers. Our Russian domain classifier used RuBERT Base \cite{kuratov2019adaptation}. An equal amount of 45K negative and positive classification examples were collected from the parallel En $\leftrightarrow$ Ru WMT'21 news task training data and the WMT'20 Biomedical Shared Task train set respectively. 

We classified the English half of the entire ~26M parallel En $\leftrightarrow$ Ru WMT'21 news task training data, saving all sentences with predicted biomedical domain probabilities over 50\%. We then used our Russian classifier to predict biomedical domain probabilities for the Russian half of the parallel data. We averaged the classifier scores from the English and Russian domain classifiers and used this averaged score as our final selection criteria. See Table \ref{mind-parallel-data-results} for data totals corresponding to different probability score cutoffs.

\subsection{Monolingual Data} \label{mono-data}
We trained consumer electronic and medical domain binary classifiers to select in-domain monolingual data from the cc100 dataset \citep{conneau-etal-2020-unsupervised, wenzek-etal-2020-ccnet} \footnote{http://data.statmt.org/cc-100/}. When training the classifiers, target side in-domain data was used for the positive class and an equal amount of randomly sampled cc100 data was collected for the negative. After a total of 500k English sentences were classified as in-domain, the top 200k, 50k and \textit{n} (where \textit{n} is commensurate with parallel data totals for a given domain) examples with the highest in-domain probabilities were used in experiments.

\section{Experimental Setup}

\subsection{Base Models}
We start by training strong baseline models for all four language pairs: Spanish, Chinese, Russian and Japaneses to English. We train our models on WMT'21 news data. Table \ref{baselines} shows initial SacreBLEU \cite{post2018} results of our models on WMT'20 test sets as well as in-domain test sets. Our models are based on the transformer large architecture \cite{vaswani2017}. As suggested in \citet{Shoeybi2019MegatronLMTM}, we move the layer normalization step for every transformer block to before each multi-head attention and feed forward sub-layer. The NMT models have 240M parameters. They took between 22 and 24 hours to train on 64 Tesla-V100 32GB GPUs with a per GPU batch size of 16k tokens. We use an initial learning rate between 1e-4 and 5e-4 with between 8k and 30k warm-up steps and an Adam \cite{Kingma2015AdamAM} optimizer. 

\begin{table}
\centering
\resizebox{\columnwidth}{!}{
\begin{tabular}{ccccc}
\hline
\textbf{Language pair} & \textbf{WMT} & \textbf{CE} & \textbf{Medical} & \textbf{Biomed}\\
\hline
Zh $\rightarrow$ En & 24.5  & 34.5 & 29.9 & - \\
Ja $\rightarrow$ En & 19.8  & 36.1 & 26.8 & - \\
Es $\rightarrow$ En & 39.9 & 46.1 & 50.1 & - \\
Ru $\rightarrow$ En & 36.2  & 25.6 & 27.7 & 38.5 \\
\hline\end{tabular}}
\caption{\label{baselines}
SacreBLEU scores of baseline models on WMT'20 for all language pairs except Es $\rightarrow$ En, and in-domain test sets for all languages. The Es $\rightarrow$ En scores are on WMT'12.}
\end{table}

We use byte-pair encoding (BPE) \cite{sennrich-etal-2016-neural} to create our NMT vocabularies. The BPE model was trained using the original domain data. The Zh $\rightarrow$ En, Ja $\rightarrow$ En, and Ru $\rightarrow$ En translation models have separate encoder and decoder vocabularies, while our Es $\rightarrow$ En model shares a single vocabulary between the encoder and decoder. Each vocabulary has 32k tokens. Our reverse direction base models (En $\rightarrow$ Other Language) used for back-translation experiments were trained in the same manner and with the same transformer architecture as our baseline forward direction models. 

\subsection{Language Models}
Our language models use a similar 16-layer transformer decoder architecture to \citet{radford2019language} with the same pre-layer normalization edit recommended by \citet{Shoeybi2019MegatronLMTM} as in our base NMT models. Though all the language models are English, they are each distinctly trained for every language pair to ensure the decoder and language models have the same tokenizer vocabulary. They are all trained on News Crawl \footnote{http://data.statmt.org/news-crawl/} English data, then fine-tuned on the English half of the in-domain parallel datasets separately.

\subsection{Adaptation}
We fixed the fine-tuning learning rates to be between 1e-5 and 5e-6. Models were fine-tuned on 1 Tesla-V100 16GB GPU until in-domain validation BLEU scores plateaued. BLEU plateau occurred after only 1 epoch for Es-En fine-tuning experiments with a batch size of 1024 tokens. Zh-En, Ja-En, Ru-En models' validation BLEU stopped improving after 15-20 epochs, while the Ru-En models for the biomedical domain finished training after 1 epoch. 

We back-translate our monolingual data described in section \ref{mono-data} with our reverse direction models generating synthetic parallel data from the top 200k, top 50k, and top $n$ (where $n$ equals the number parallel examples for that language pair and domain) monolingual examples. The top $n$ and top 50k parallel examples are a higher quality subset of the 200k examples, allowing us to characterize the impact of quantity verses quality of back-translated data in a low resource environment. We fine-tune our base models exclusively on back-translated data for our target side monolingual experiments and on a mix of human-translated and back-translated data for our combined parallel and target monolingual experiments.

\section{In-Domain Parallel Results}
The in-domain parallel results averaged across all language pairs and across the consumer electronic and medical domains are displayed in Table \ref{mono-only-and-parallel-only-results-summary}. Mixed-domain fine-tuning has slightly lower in-domain performance on average (a -0.6 difference) compared with fine-tuning on in-domain parallel data only. Mixed domain fine-tuning does help maintain original domain performance. We see an average original domain BLEU score of 27.4 for models tuned on mixed domain data and an average score of 23.8 for those fine-tuned on in-domain data only. Shallow fusion decoding with an in-domain language model boosts performance for all languages and domains.

\begin{table*}
    \centering
    \begin{tabular}{clcc}
    \hline
    \textbf{Scenario}  & \textbf{Adaptation Method} & \textbf{New-Domain Avg.} &  \textbf{Original Domain Avg.} \\
    \hline
    \multirow{5}{*}{Parallel Only} & Baseline & 34.6 & 30.1 \\
                                   & Ensemble & 39.3 & \bf{29.4} \\
                                   & Mixed-Domain & 42.1 & 27.4 \\
                                   & In-Domain & 42.7 & 23.8 \\
                                   & In-Domain + SF & \bf{43.0} & 25.8 \\
    \hline
    \multirow{5}{*}{Target Monolingual Only} & BT Top200k cc100 & 35.4 & 28.4 \\
                                             & BT Top50k cc100 & 35.9 & 27.3 \\
                                             & BT TopN cc100 & 35.8 & \bf{28.9} \\
                                             & BT Target Half & 39.0 & 27.3 \\
                                             & BT Target Half + SF & \bf{40.0} & 27.2 \\
    \hline
    \end{tabular}
    \caption{\label{mono-only-and-parallel-only-results-summary}
    A summary of results for parallel only and target monolingual only experiments. Each value is the BLEU score for the corresponding adaptation method averaged over every language pair and over the consumer electronic and medical domains. Ensemble denotes the scores for an ensemble of the baseline model and models fine-tuned on in-domain data. SF stands for shallow fusion. BT stands for back translation. TopN models were fine-tuned on an amount of back-translated cc100 data equal to the number of examples in their corresponding in-domain datasets.}
\end{table*}

\subsection{Mitigating Original Domain Degradation via Ensembling}
\begin{figure*}[htb!]
    \centering
    \begin{subfigure}[b]{0.45\textwidth}
        \centering
        \includegraphics[width=\textwidth]{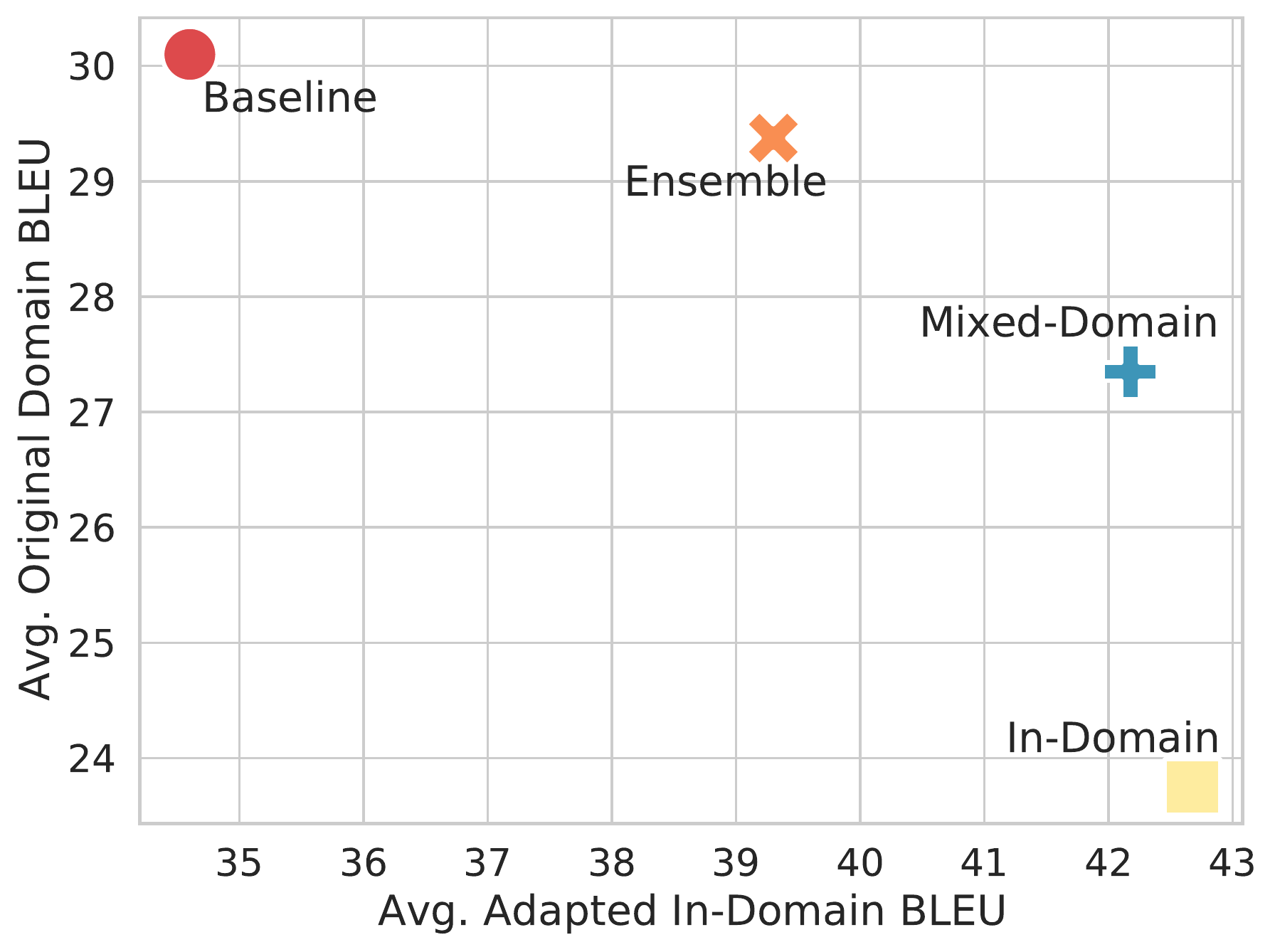}
        \caption{in-domain averages}
        \label{fig:in-domain-avg}
    \end{subfigure}
    \hfill
    \begin{subfigure}[b]{0.45\textwidth}
        \centering
        \includegraphics[width=\textwidth]{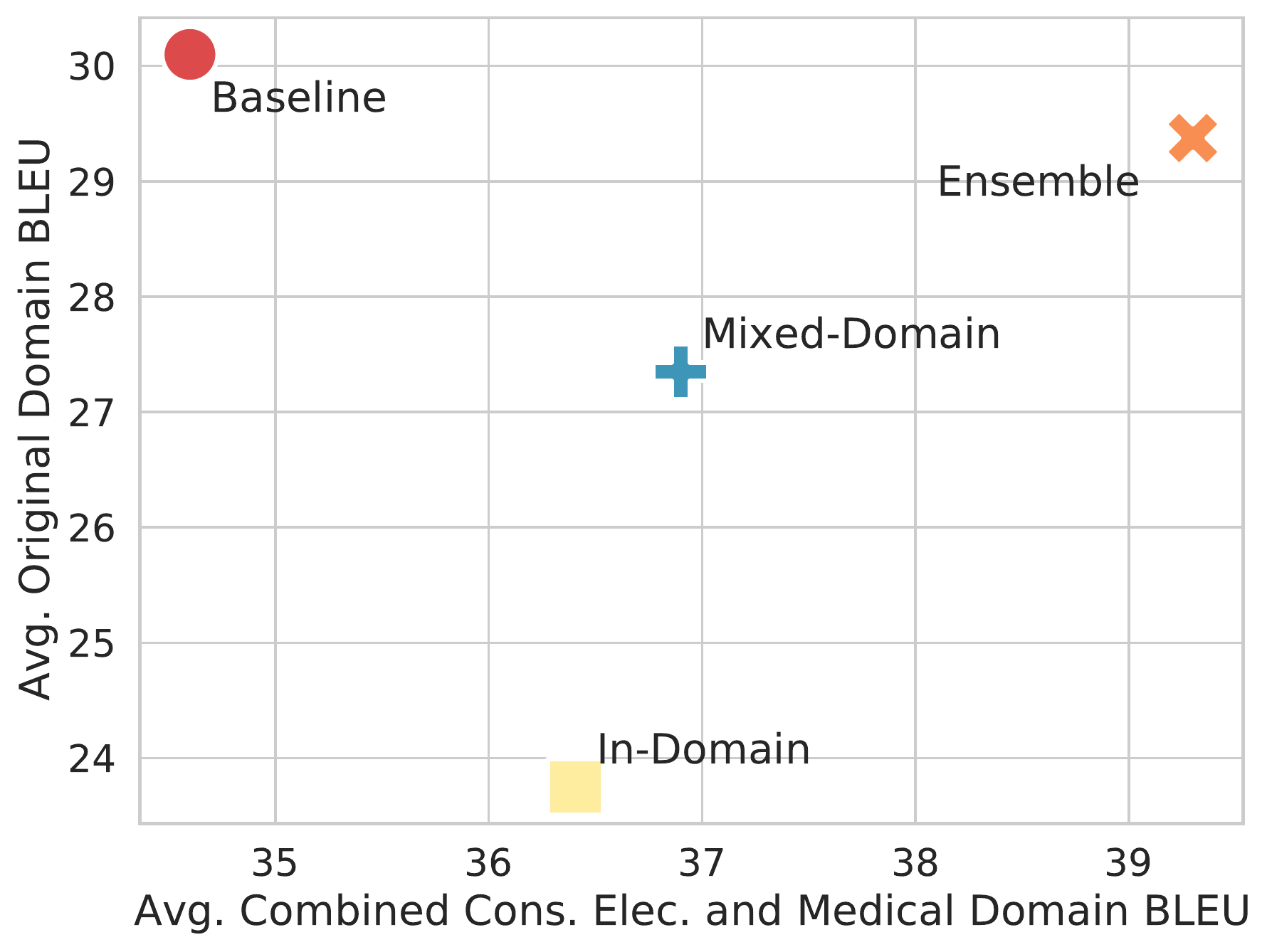}
        \caption{in-and-out of domain averages}
        \label{fig:in-and-out-of-domain-avg}
    \end{subfigure}
    \caption{Original vs. new domain performance trade-off across parallel adaptation methods. (a) shows the average original domain performance as a function of the average \textit{in-domain} BLEU score for each new domain across all languages, capturing this trade-off when translating one new domain at a time.  (b) displays the average \textit{in-and-out of domain} BLEU scores for each adaptation method over all language pairs, encapsulating trade off trends when translating text from multiple new domains simultaneously.}
    \label{fig:parallel-method-results-and-tradeoffs}
\end{figure*}

We ensemble our fine-tuned in-domain parallel and baseline models together. When ensembled, baseline performance remains within 0.7 BLEU of its original score across all languages. This is a huge improvement over the 10+ BLEU score drop seen when fine-tuning on the consumer electronic domain. No ensemble out performs their single fine-tuned model counterparts when evaluated on in-domain data. Nevertheless, the ensemble still achieves a several BLEU point improvement in each domain over the baseline and the average BLEU score across all domains is much higher when additionally comparing against any single model's out-of-domain performance. These results indicate that, when translating mixed domain or unknown domain data, ensembling in-domain models should lead to higher quality translations-- even when domains are drastically different (e.g. the consumer electronic and medical domains). Figure \ref{fig:parallel-method-results-and-tradeoffs} presents the original vs. new domain trade-off for the consumer electronic and medical domains averaged over all language pairs. Figure \ref{fig:in-and-out-of-domain-avg} highlights the advantage of ensembling. The x-axis values in \ref{fig:in-and-out-of-domain-avg} are the combined average consumer electronic and medical domain BLEU scores irrespective of the domain for which each model was fine-tuned. 

\subsection{Benefits of Mined In-Domain Parallel Data}
Fine-tuning the baseline Ru $\rightarrow$ En model with combined mined and original parallel data increased performance over fine-tuning with original data alone by 0.2 and 0.7 BLEU. A higher domain probability cutoff threshold, favoring reduced in-domain noise over larger data quantity, resulted in a 0.5 BLEU score difference between the two models trained with mined data. It should be noted that the additional parallel data was mined from the parallel Ru $\rightarrow$ En training set used to train the baseline model. Though the model saw all mined examples during initial baseline training, viewing these in-domain examples again during the fine-tuning stage still increased in-domain performance over fine-tuning on purely unseen data. See Table \ref{mind-parallel-data-results} for a result breakdown.

\begin{table}
\centering
\resizebox{\columnwidth}{!}{
\begin{tabular}{lccc}
\hline
\textbf{Model Description} & \textbf{Cutoff} & \textbf{Total} & \textbf{BLEU} \\
\hline
Baseline & - & - & 38.5 \\

Original Parallel & - & 46,782 & 41.3 \\

Original Parallel + Mined & .90 & 254,037 & 41.5 \\

\textbf{Original Parallel + Mined} & \textbf{.97} & \textbf{140,414} & \textbf{42.0} \\
\hline
\end{tabular}}
\caption{\label{mind-parallel-data-results}
The performance increase from adding mined parallel data to the biomedical Ru $\rightarrow$ En fine-tuning set. "cutoff" is the domain classifier probability threshold and "total" is the train set size with mined examples added. 
}
\end{table}

\begin{figure*}
    \centering
    \begin{subfigure}[b]{0.32\textwidth}
        \centering
        \includegraphics[width=\textwidth]{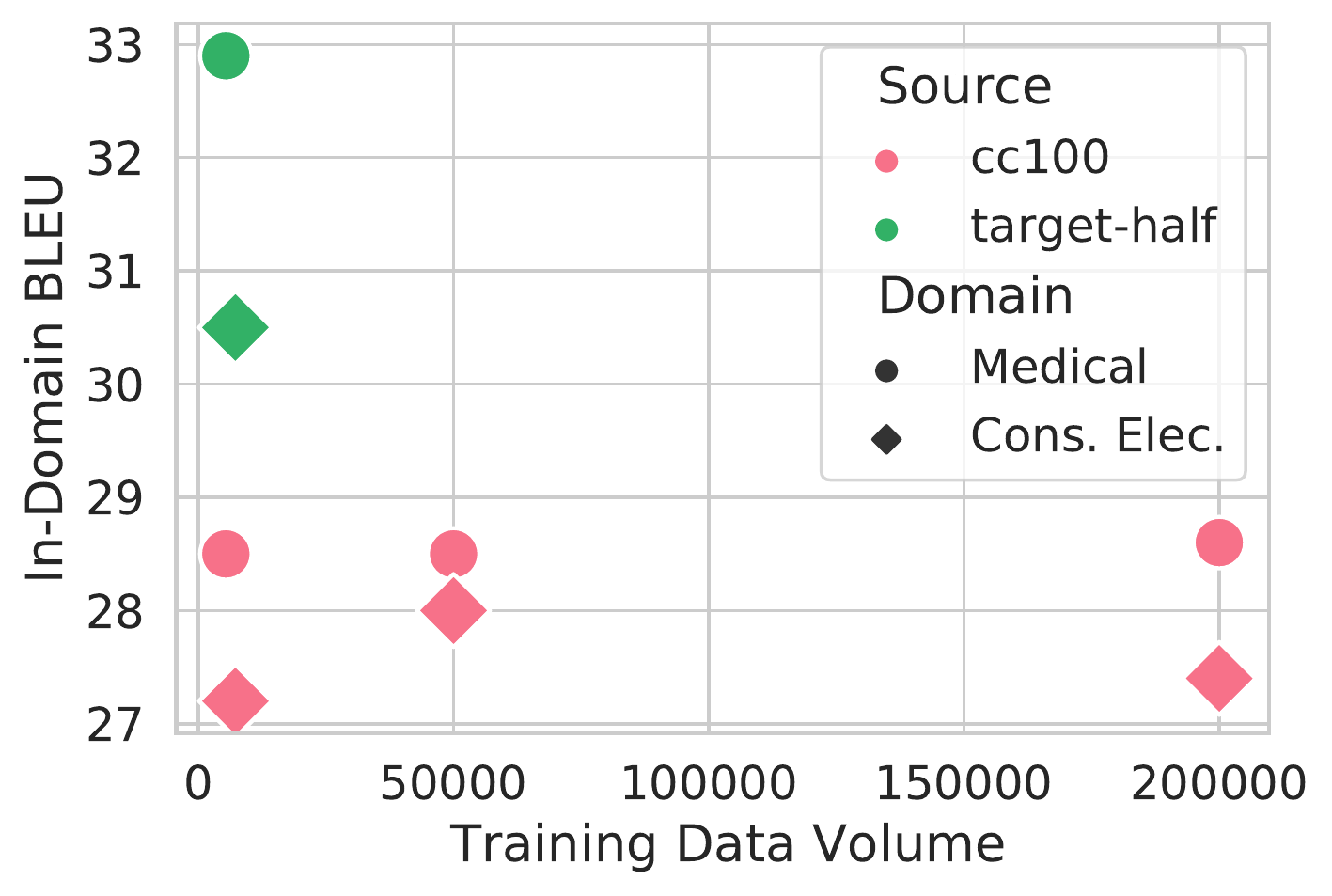}
        \caption{Zh $\rightarrow$ En}
        \label{fig:quant-qual-ru}
    \end{subfigure}
    \hfill
    \begin{subfigure}[b]{0.32\textwidth}
        \centering
        \includegraphics[width=\textwidth]{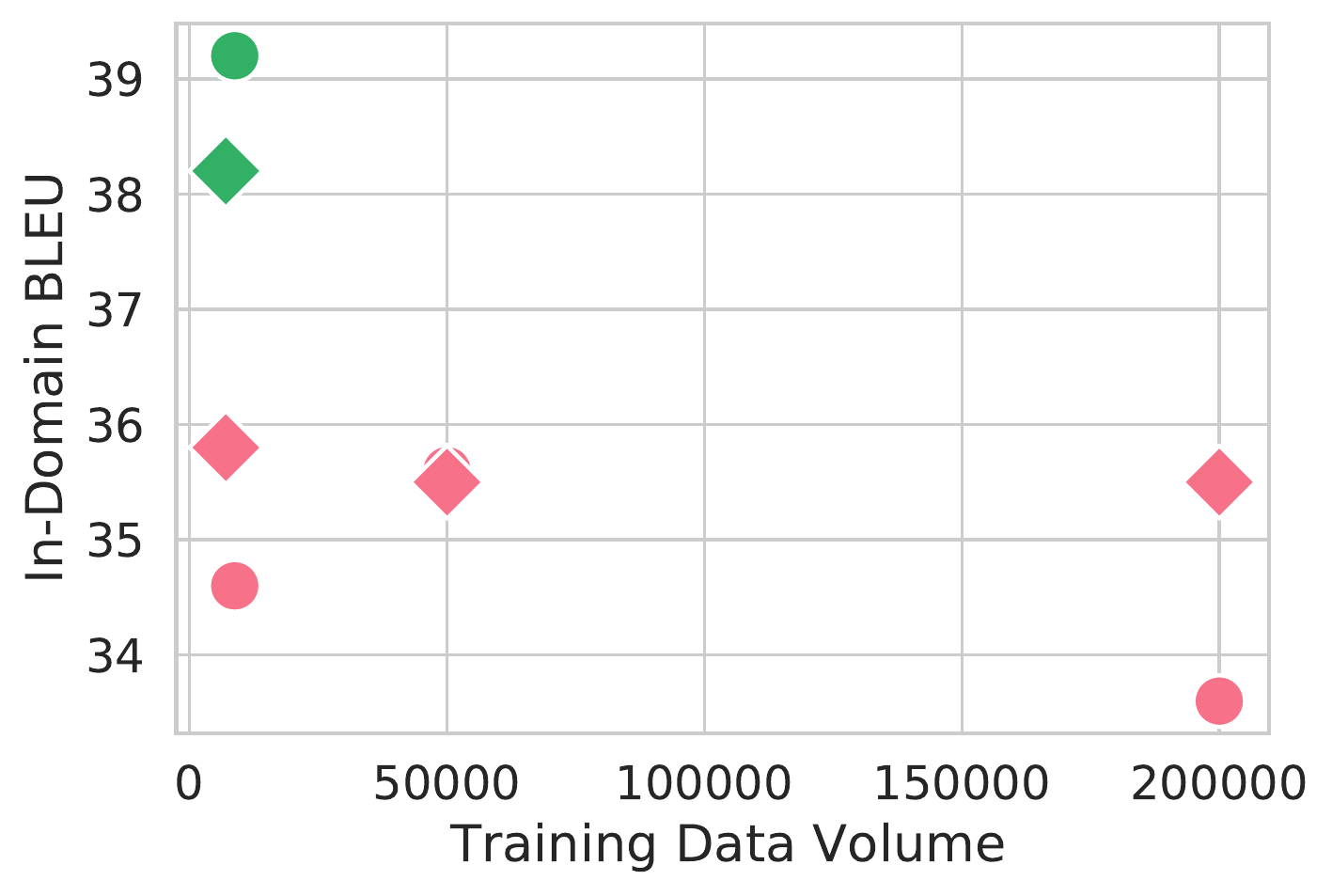}
        \caption{Ru $\rightarrow$ En}
        \label{fig:quant-qual-zh}
    \end{subfigure}
    \hfill
    \begin{subfigure}[b]{0.32\textwidth}
        \centering
        \includegraphics[width=\textwidth]{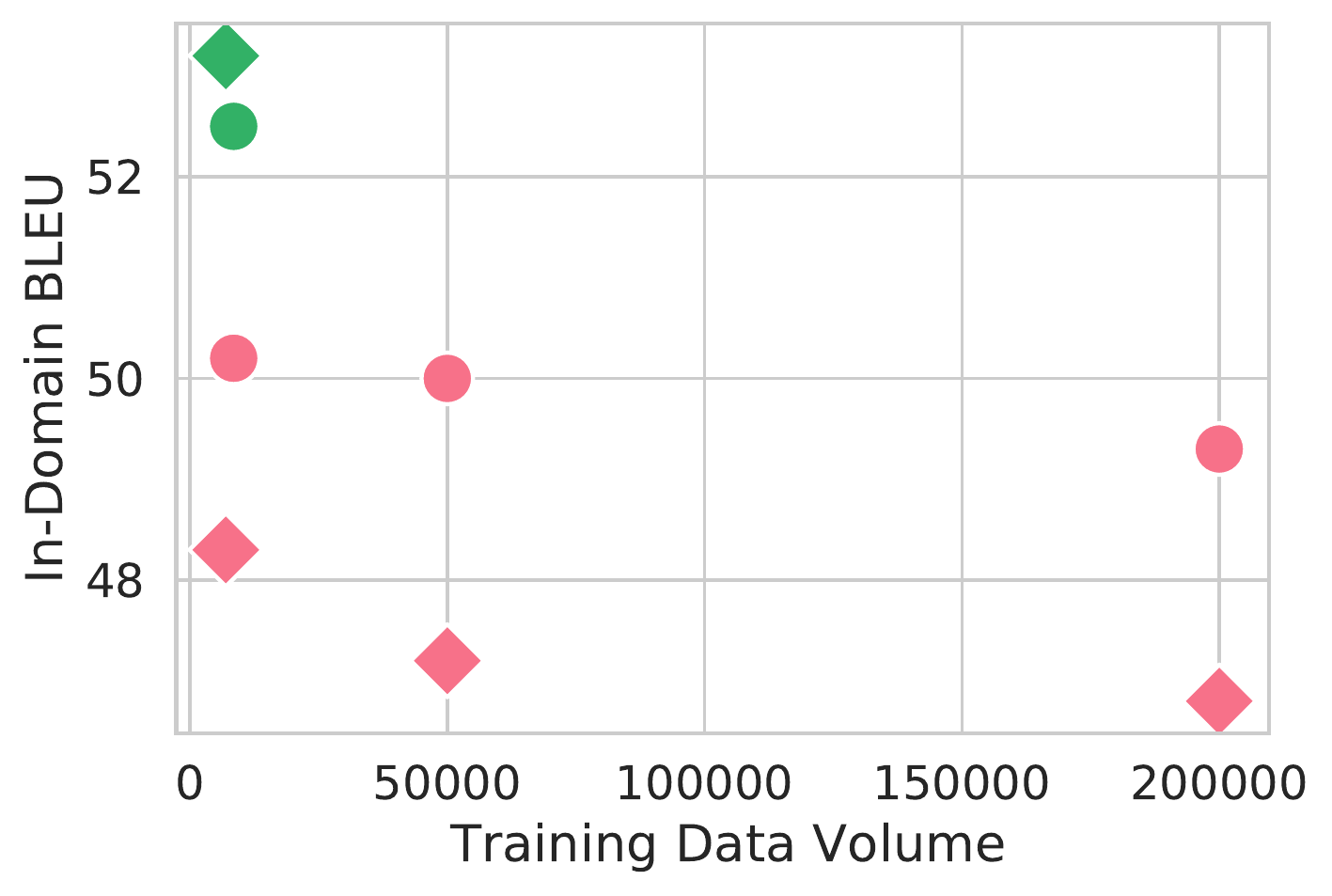}
        \caption{Es $\rightarrow$ En}
        \label{fig:quant-qual-es}
    \end{subfigure}
    \caption{In-Domain BLEU scores after fine-tuning the baseline model on back-translated data. The green points correspond to scores from models fine-tuned on the back-translated target-half of the in-domain parallel datasets. The pink points are from models fine-tuned on back-translated cc100 data. Models with scores shown in green saw smaller volumes of high quality synthetic data compared to those in pink.}
    \label{fig:backtrans-quantity-vs-quality}
\end{figure*}

\section{Target Side Monolingual Results}
In the bottom half of Table \ref{mono-only-and-parallel-only-results-summary} we see that fine-tuning a base model on high quality back-translated data comes within approximately 3 BLEU points of fine-tuning on human translated in-domain parallel data on average. When analyzing individual models, the best Ja $\rightarrow$ En monolingual model matched the performance of the in-domain parallel model for the medical domain and surpassed it by 0.7 BLEU points in the consumer electronic domain.

\subsection{Back-Translated Quantity vs. Quality Trade-Off}
We compare fine-tuning on back-translated data mined from cc100 with fine-tuning on the back-translated English half of each in-domain parallel dataset. Across the language pairs, there seems to be no major difference in performance between models fine-tuned with 200k, 50k, or top \textit{n} totals of back-translated cc100 data. They each reach an average BLEU score between 35 and 36 as seen in Table \ref{mono-only-and-parallel-only-results-summary}. When base models are fine-tuned on the back-translated target half of the original in-domain parallel datasets, the model's performance increased by an average of 3.3 BLEU compared to the cc100 back-translation experiments. Even with over 20x less data, fine-tuning on clean (in terms of domain accuracy) back-translated examples out scores utilizing noisier data. This point is also illustrated in Figure \ref{fig:backtrans-quantity-vs-quality}. 

\subsection{Shallow Fusion}
Across the board shallow fusion leads to within 1.0 BLEU score increase compared to the baseline scores in each domain. For Ru $\rightarrow$ En, Es $\rightarrow$ En, and Ja $\rightarrow$ En shallow fusion with in-domain language models also increases \textit{original domain} performance within 1.0 BLEU point of their original WMT'20 scores. This shows that even language models fine-tuned on out of domain data still have an advantageous impact when used for shallow fusion decoding. 

\section{In-Domain Parallel + Target Side Monolingual Results}
We experimented with a number of approaches to combining back-translated data with in-domain parallel data. A summary of results for these experiments can be seen in Table \ref{mono-and-parallel-combined-results-summary}. We first used our baseline reverse direction model to back-translate the top 50k cc100 sentences from each domain. Baseline models fine-tuned on a mix of this data and in-domain parallel data improved an average of 6.2 BLEU points from the baseline. We then tried fine-tuning our \textit{reverse} direction model on our parallel domain data before back-translation. Combining this back-translated data with human translated parallel-data resulted in another +0.7 BLEU increase on average. Next we experimented with tagged back-translation. We prepended a special back-translation token ($<BT>$) to the beginning of every synthetic back-translated input from our previous iteration. Tagging back-translated examples actually slightly decreased the average BLEU score by -0.1 compared to not adding tags. Finally, we used in-domain shallow fusion decoding at inference time with our model fine-tuned via tagged back-translation for a +0.6 average performance boost. Despite our efforts, we found none to be as effective as fine-tuning on purely in-domain data or a mix of in-domain and out-of-domain parallel data.

\begin{table}
    \centering
    \resizebox{\columnwidth}{!}{
    \begin{tabular}{lc}
    \hline
    \textbf{Method} & \textbf{BLEU} \\
    \hline
    Baseline & 30.6\\
    In-Domain Parallel + BT w/ Base & 36.8 \\
    In-Domain Parallel + BT w/ Tuned & 37.5 \\
    In-Domain Parallel + Tagged BT w/ Tuned & 37.4 \\
    In-Domain Parallel + Tagged BT w/ Tuned + SF & 38.0 \\
    \hline
    Mixed-Domain Parallel & 38.7 \\
    In-Domain Parallel Only & 38.9 \\
    In-Domain Parallel Only + SF & \bf{39.0} \\
    \hline
    \end{tabular}}
    \caption{\label{mono-and-parallel-combined-results-summary}
    In-domain parallel + target side monolingual results for Ru$\rightarrow$En BLEU scores averaged across the consumer electronic, medical, and biomedical domains. Base models were fine-tuned on a mix of in-domain parallel data and back-translated Top50k cc100 data. SF stands for shallow fusion and BT stand for back-translation. "w/ Tuned" indicates where target side monolingual data was back-translated with a reverse direction model that has been fine-tuned on parallel in-domain data. Methods using human translated parallel data alone out preformed those combining back-translated and human translated parallel data.}
\end{table}

\section{Recommendations}
\begin{enumerate}
    \item In low resource situations, with access to both parallel and monolingual data (<200k monolingual examples, <10k parallel examples), don't spend time on back-translation. Instead focus on parallel in-domain and mixed domain fine-tuning. 
    \item Ensemble in-domain and baseline models for more robust translations when translating mixed or unknown domains.
     \item Use an in-domain language model for shallow fusion decoding. It will most likely improve both your in-domain and original domain performance, especially when parallel domain data is not available. In-domain shallow fusion can be an effective adaptation approach even without fine-tuning the baseline translation model.
    \item If you only have monolingual data, back-translate the highest quality monolingual data possible, prioritize quality over data volume in low resource settings (<200k monolingual examples).
    \item It is better to mine a moderate amount of parallel data over a larger amount of in-domain monolingual data.
\end{enumerate}

\section{Conclusion}
We conduct an empirical study comparing parallel and monolingual data approaches to domain adaptation in NMT. We made recommendations on how to achieve the best in-domain translation performance with access to low resource parallel and/or monolingual domain data. Additionally, we explored model ensembling to reduce regression of original domain performance and the benefits of mined in-domain parallel data. We hope this work can guide others in their creation of high quality domain specific machine translation systems.

\bibliography{custom}

\begin{thebibliography}{36}
\expandafter\ifx\csname natexlab\endcsname\relax\def\natexlab#1{#1}\fi

\bibitem[{Ahmadnia and Dorr(2019)}]{AhmadniaDorr+2019+268+278}
Benyamin Ahmadnia and Bonnie~J. Dorr. 2019.
\newblock \href {https://doi.org/doi:10.1515/comp-2019-0019} {Augmenting neural
  machine translation through round-trip training approach}.
\newblock \emph{Open Computer Science}, 9(1):268--278.

\bibitem[{Artetxe et~al.(2018)Artetxe, Labaka, and
  Agirre}]{artetxe-etal-2018-unsupervised}
Mikel Artetxe, Gorka Labaka, and Eneko Agirre. 2018.
\newblock \href {https://doi.org/10.18653/v1/D18-1399} {Unsupervised
  statistical machine translation}.
\newblock In \emph{Proceedings of the 2018 Conference on Empirical Methods in
  Natural Language Processing}, pages 3632--3642, Brussels, Belgium.
  Association for Computational Linguistics.

\bibitem[{Artetxe and Schwenk(2019)}]{artetxe-schwenk-2019-massively}
Mikel Artetxe and Holger Schwenk. 2019.
\newblock \href {https://doi.org/10.1162/tacl_a_00288} {Massively multilingual
  sentence embeddings for zero-shot cross-lingual transfer and beyond}.
\newblock \emph{Transactions of the Association for Computational Linguistics},
  7:597--610.

\bibitem[{Caswell et~al.(2019)Caswell, Chelba, and
  Grangier}]{caswell-etal-2019-tagged}
Isaac Caswell, Ciprian Chelba, and David Grangier. 2019.
\newblock \href {https://doi.org/10.18653/v1/W19-5206} {Tagged
  back-translation}.
\newblock In \emph{Proceedings of the Fourth Conference on Machine Translation
  (Volume 1: Research Papers)}, pages 53--63, Florence, Italy. Association for
  Computational Linguistics.

\bibitem[{Chu et~al.(2017)Chu, Dabre, and Kurohashi}]{chu-etal-2017-empirical}
Chenhui Chu, Raj Dabre, and Sadao Kurohashi. 2017.
\newblock \href {https://doi.org/10.18653/v1/P17-2061} {An empirical comparison
  of domain adaptation methods for neural machine translation}.
\newblock In \emph{Proceedings of the 55th Annual Meeting of the Association
  for Computational Linguistics (Volume 2: Short Papers)}, pages 385--391,
  Vancouver, Canada. Association for Computational Linguistics.

\bibitem[{Chu et~al.(2018)Chu, Dabre, and Kurohashi}]{Chenhui-Chu2018}
Chenhui Chu, Raj Dabre, and Sadao Kurohashi. 2018.
\newblock \href {https://doi.org/10.2197/ipsjjip.26.529} {A comprehensive
  empirical comparison of domain adaptation methods for neural machine
  translation}.
\newblock \emph{Journal of Information Processing}, 26:529--538.

\bibitem[{Chu and Wang(2018)}]{chu-wang-2018-survey}
Chenhui Chu and Rui Wang. 2018.
\newblock \href {https://aclanthology.org/C18-1111} {A survey of domain
  adaptation for neural machine translation}.
\newblock In \emph{Proceedings of the 27th International Conference on
  Computational Linguistics}, pages 1304--1319, Santa Fe, New Mexico, USA.
  Association for Computational Linguistics.

\bibitem[{Conneau et~al.(2020)Conneau, Khandelwal, Goyal, Chaudhary, Wenzek,
  Guzm{\'a}n, Grave, Ott, Zettlemoyer, and
  Stoyanov}]{conneau-etal-2020-unsupervised}
Alexis Conneau, Kartikay Khandelwal, Naman Goyal, Vishrav Chaudhary, Guillaume
  Wenzek, Francisco Guzm{\'a}n, Edouard Grave, Myle Ott, Luke Zettlemoyer, and
  Veselin Stoyanov. 2020.
\newblock \href {https://doi.org/10.18653/v1/2020.acl-main.747} {Unsupervised
  cross-lingual representation learning at scale}.
\newblock In \emph{Proceedings of the 58th Annual Meeting of the Association
  for Computational Linguistics}, pages 8440--8451, Online. Association for
  Computational Linguistics.

\bibitem[{Conneau and Lample(2019)}]{NEURIPS2019_c04c19c2}
Alexis Conneau and Guillaume Lample. 2019.
\newblock \href
  {https://proceedings.neurips.cc/paper/2019/file/c04c19c2c2474dbf5f7ac4372c5b9af1-Paper.pdf}
  {Cross-lingual language model pretraining}.
\newblock In \emph{Advances in Neural Information Processing Systems},
  volume~32. Curran Associates, Inc.

\bibitem[{Devlin et~al.(2018)Devlin, Chang, Lee, and
  Toutanova}]{devlin2018bert}
Jacob Devlin, Ming-Wei Chang, Kenton Lee, and Kristina Toutanova. 2018.
\newblock Bert: Pre-training of deep bidirectional transformers for language
  understanding.
\newblock \emph{arXiv preprint arXiv:1810.04805}.

\bibitem[{Dou et~al.(2019{\natexlab{a}})Dou, Wang, Hu, and
  Neubig}]{dou2019domain}
Zi-Yi Dou, Xinyi Wang, Junjie Hu, and Graham Neubig. 2019{\natexlab{a}}.
\newblock \href {http://arxiv.org/abs/1910.02555} {Domain differential
  adaptation for neural machine translation}.

\bibitem[{Dou et~al.(2019{\natexlab{b}})Dou, Wang, Hu, and
  Neubig}]{DBLP:journals/corr/abs-1910-02555}
Zi{-}Yi Dou, Xinyi Wang, Junjie Hu, and Graham Neubig. 2019{\natexlab{b}}.
\newblock \href {http://arxiv.org/abs/1910.02555} {Domain differential
  adaptation for neural machine translation}.
\newblock \emph{CoRR}, abs/1910.02555.

\bibitem[{Edunov et~al.(2018)Edunov, Ott, Auli, and
  Grangier}]{edunov2018understanding}
Sergey Edunov, Myle Ott, Michael Auli, and David Grangier. 2018.
\newblock \href {http://arxiv.org/abs/1808.09381} {Understanding
  back-translation at scale}.

\bibitem[{Gu et~al.(2018)Gu, Wang, Chen, Li, and Cho}]{gu-etal-2018-meta}
Jiatao Gu, Yong Wang, Yun Chen, Victor O.~K. Li, and Kyunghyun Cho. 2018.
\newblock \href {https://doi.org/10.18653/v1/D18-1398} {Meta-learning for
  low-resource neural machine translation}.
\newblock In \emph{Proceedings of the 2018 Conference on Empirical Methods in
  Natural Language Processing}, pages 3622--3631, Brussels, Belgium.
  Association for Computational Linguistics.

\bibitem[{Gulcehre et~al.(2015)Gulcehre, Firat, Xu, Cho, Barrault, Lin,
  Bougares, Schwenk, and Bengio}]{gulcehre2015using}
Caglar Gulcehre, Orhan Firat, Kelvin Xu, Kyunghyun Cho, Loic Barrault, Huei-Chi
  Lin, Fethi Bougares, Holger Schwenk, and Yoshua Bengio. 2015.
\newblock On using monolingual corpora in neural machine translation.
\newblock \emph{arXiv preprint arXiv:1503.03535}.

\bibitem[{He et~al.(2016)He, Xia, Qin, Wang, Yu, Liu, and
  Ma}]{NIPS2016_5b69b9cb}
Di~He, Yingce Xia, Tao Qin, Liwei Wang, Nenghai Yu, Tie-Yan Liu, and Wei-Ying
  Ma. 2016.
\newblock \href
  {https://proceedings.neurips.cc/paper/2016/file/5b69b9cb83065d403869739ae7f0995e-Paper.pdf}
  {Dual learning for machine translation}.
\newblock In \emph{Advances in Neural Information Processing Systems},
  volume~29. Curran Associates, Inc.

\bibitem[{Houlsby et~al.(2019)Houlsby, Giurgiu, Jastrzebski, Morrone,
  De~Laroussilhe, Gesmundo, Attariyan, and Gelly}]{houlsby2019parameter}
Neil Houlsby, Andrei Giurgiu, Stanislaw Jastrzebski, Bruna Morrone, Quentin
  De~Laroussilhe, Andrea Gesmundo, Mona Attariyan, and Sylvain Gelly. 2019.
\newblock Parameter-efficient transfer learning for nlp.
\newblock In \emph{International Conference on Machine Learning}, pages
  2790--2799. PMLR.

\bibitem[{Kingma and Ba(2015)}]{Kingma2015AdamAM}
Diederik~P. Kingma and Jimmy Ba. 2015.
\newblock Adam: A method for stochastic optimization.
\newblock \emph{CoRR}, abs/1412.6980.

\bibitem[{Kumari et~al.(2021)Kumari, Jaiswal, Patidar, Patwardhan, Karande,
  Agarwal, and Vig}]{kumari-etal-2021-domain}
Surabhi Kumari, Nikhil Jaiswal, Mayur Patidar, Manasi Patwardhan, Shirish
  Karande, Puneet Agarwal, and Lovekesh Vig. 2021.
\newblock \href {https://aclanthology.org/2021.adaptnlp-1.26} {Domain
  adaptation for {NMT} via filtered iterative back-translation}.
\newblock In \emph{Proceedings of the Second Workshop on Domain Adaptation for
  NLP}, pages 263--271, Kyiv, Ukraine. Association for Computational
  Linguistics.

\bibitem[{Kuratov and Arkhipov(2019)}]{kuratov2019adaptation}
Yuri Kuratov and Mikhail Arkhipov. 2019.
\newblock Adaptation of deep bidirectional multilingual transformers for
  russian language.
\newblock \emph{arXiv preprint arXiv:1905.07213}.

\bibitem[{Lample et~al.(2018)Lample, Conneau, Denoyer, and
  Ranzato}]{lample2018unsupervised}
Guillaume Lample, Alexis Conneau, Ludovic Denoyer, and Marc'Aurelio Ranzato.
  2018.
\newblock \href {https://openreview.net/forum?id=rkYTTf-AZ} {Unsupervised
  machine translation using monolingual corpora only}.
\newblock In \emph{International Conference on Learning Representations}.

\bibitem[{Post(2018)}]{post2018}
Matt Post. 2018.
\newblock A call for clarity in reporting bleu scores.
\newblock \emph{arXiv e-prints arXiv:1804.0877}.

\bibitem[{Radford et~al.(2019)Radford, Wu, Child, Luan, Amodei, Sutskever
  et~al.}]{radford2019language}
Alec Radford, Jeffrey Wu, Rewon Child, David Luan, Dario Amodei, Ilya
  Sutskever, et~al. 2019.
\newblock Language models are unsupervised multitask learners.
\newblock \emph{OpenAI blog}, 1(8):9.

\bibitem[{Saunders(2021)}]{saunders2021domain}
Danielle Saunders. 2021.
\newblock \href {http://arxiv.org/abs/2104.06951} {Domain adaptation and
  multi-domain adaptation for neural machine translation: A survey}.

\bibitem[{Sennrich et~al.(2016{\natexlab{a}})Sennrich, Haddow, and
  Birch}]{sennrich-etal-2016-improving}
Rico Sennrich, Barry Haddow, and Alexandra Birch. 2016{\natexlab{a}}.
\newblock \href {https://doi.org/10.18653/v1/P16-1009} {Improving neural
  machine translation models with monolingual data}.
\newblock In \emph{Proceedings of the 54th Annual Meeting of the Association
  for Computational Linguistics (Volume 1: Long Papers)}, pages 86--96, Berlin,
  Germany. Association for Computational Linguistics.

\bibitem[{Sennrich et~al.(2016{\natexlab{b}})Sennrich, Haddow, and
  Birch}]{sennrich-etal-2016-neural}
Rico Sennrich, Barry Haddow, and Alexandra Birch. 2016{\natexlab{b}}.
\newblock \href {https://doi.org/10.18653/v1/P16-1162} {Neural machine
  translation of rare words with subword units}.
\newblock In \emph{Proceedings of the 54th Annual Meeting of the Association
  for Computational Linguistics (Volume 1: Long Papers)}, pages 1715--1725,
  Berlin, Germany. Association for Computational Linguistics.

\bibitem[{Shoeybi et~al.(2019)Shoeybi, Patwary, Puri, LeGresley, Casper, and
  Catanzaro}]{Shoeybi2019MegatronLMTM}
Mohammad Shoeybi, Mostofa~Ali Patwary, Raul Puri, Patrick LeGresley, Jared
  Casper, and Bryan Catanzaro. 2019.
\newblock Megatron-lm: Training multi-billion parameter language models using
  model parallelism.
\newblock \emph{ArXiv}, abs/1909.08053.

\bibitem[{Song et~al.(2019)Song, Tan, Qin, Lu, and Liu}]{song2019mass}
Kaitao Song, Xu~Tan, Tao Qin, Jianfeng Lu, and Tie-Yan Liu. 2019.
\newblock Mass: Masked sequence to sequence pre-training for language
  generation.
\newblock \emph{ICML 2019}.

\bibitem[{Sun et~al.(2019)Sun, Wang, Chen, Utiyama, Sumita, and
  Zhao}]{Sun2019AnES}
Haipeng Sun, Rui Wang, Kehai Chen, M.~Utiyama, E.~Sumita, and T.~Zhao. 2019.
\newblock An empirical study of domain adaptation for unsupervised neural
  machine translation.
\newblock \emph{ArXiv}, abs/1908.09605.

\bibitem[{Thompson and Koehn(2019)}]{thompson-koehn-2019-vecalign}
Brian Thompson and Philipp Koehn. 2019.
\newblock \href {https://doi.org/10.18653/v1/D19-1136} {{V}ecalign: Improved
  sentence alignment in linear time and space}.
\newblock In \emph{Proceedings of the 2019 Conference on Empirical Methods in
  Natural Language Processing and the 9th International Joint Conference on
  Natural Language Processing (EMNLP-IJCNLP)}, pages 1342--1348, Hong Kong,
  China. Association for Computational Linguistics.

\bibitem[{Vaswani et~al.(2017)Vaswani, Shazeer, Parmar, Uszkoreit, Jones,
  Gomez, Kaiser, and Polosukhin}]{vaswani2017}
Ashish Vaswani, Noam Shazeer, Niki Parmar, Jakob Uszkoreit, Llion Jones,
  Aidan~N. Gomez, Lukasz Kaiser, and Illia Polosukhin. 2017.
\newblock Attention is all you need.
\newblock \emph{arXiv preprint arXiv: 1706.03762}.

\bibitem[{Wenzek et~al.(2020)Wenzek, Lachaux, Conneau, Chaudhary, Guzm{\'a}n,
  Joulin, and Grave}]{wenzek-etal-2020-ccnet}
Guillaume Wenzek, Marie-Anne Lachaux, Alexis Conneau, Vishrav Chaudhary,
  Francisco Guzm{\'a}n, Armand Joulin, and Edouard Grave. 2020.
\newblock \href {https://aclanthology.org/2020.lrec-1.494} {{CCN}et: Extracting
  high quality monolingual datasets from web crawl data}.
\newblock In \emph{Proceedings of the 12th Language Resources and Evaluation
  Conference}, pages 4003--4012, Marseille, France. European Language Resources
  Association.

\bibitem[{Wolf et~al.(2019)Wolf, Debut, Sanh, Chaumond, Delangue, Moi, Cistac,
  Rault, Louf, Funtowicz et~al.}]{wolf2019huggingface}
Thomas Wolf, Lysandre Debut, Victor Sanh, Julien Chaumond, Clement Delangue,
  Anthony Moi, Pierric Cistac, Tim Rault, R{\'e}mi Louf, Morgan Funtowicz,
  et~al. 2019.
\newblock Huggingface's transformers: State-of-the-art natural language
  processing.
\newblock \emph{arXiv preprint arXiv:1910.03771}.

\bibitem[{Zhang and Zong(2016)}]{zhang-zong-2016-exploiting}
Jiajun Zhang and Chengqing Zong. 2016.
\newblock \href {https://doi.org/10.18653/v1/D16-1160} {Exploiting source-side
  monolingual data in neural machine translation}.
\newblock In \emph{Proceedings of the 2016 Conference on Empirical Methods in
  Natural Language Processing}, pages 1535--1545, Austin, Texas. Association
  for Computational Linguistics.

\bibitem[{Zhang et~al.(2019)Zhang, Shapiro, Kumar, McNamee, Carpuat, and
  Duh}]{zhang-etal-2019-curriculum}
Xuan Zhang, Pamela Shapiro, Gaurav Kumar, Paul McNamee, Marine Carpuat, and
  Kevin Duh. 2019.
\newblock \href {https://doi.org/10.18653/v1/N19-1189} {Curriculum learning for
  domain adaptation in neural machine translation}.
\newblock In \emph{Proceedings of the 2019 Conference of the North {A}merican
  Chapter of the Association for Computational Linguistics: Human Language
  Technologies, Volume 1 (Long and Short Papers)}, pages 1903--1915,
  Minneapolis, Minnesota. Association for Computational Linguistics.

\bibitem[{Zoph et~al.(2016)Zoph, Yuret, May, and
  Knight}]{zoph-etal-2016-transfer}
Barret Zoph, Deniz Yuret, Jonathan May, and Kevin Knight. 2016.
\newblock \href {https://doi.org/10.18653/v1/D16-1163} {Transfer learning for
  low-resource neural machine translation}.
\newblock In \emph{Proceedings of the 2016 Conference on Empirical Methods in
  Natural Language Processing}, pages 1568--1575, Austin, Texas. Association
  for Computational Linguistics.

\end{thebibliography}
\bibliographystyle{acl_natbib}
\newpage

\appendix
\section{Detailed Results}
\label{sec:appendix-detailed-results}
\begin{table*}
\centering
\begin{tabular}{|c|c|l|c|c|}
\hline
\textbf{Languages}  & \textbf{Domain} & \textbf{Model Description} &  \textbf{In-Domain} & \textbf{Original Domain} \\
\hline
\multirow{10}{*}{Ja $\rightarrow$ En} & \multirow{ 4}{*}{Consumer Electronic} & Baseline & 36.1 & 19.8 \\
                                      \cline{3-5}
                                      & & Ensemble Across Domains                        & 36.5 & 20.0 \\
                                      & & Mixed-Domain Finetune                          & 37.2 & 19.4 \\
                                      & & In-Domain Finetune                             & 36.9 & 18.7 \\
                                      & & In-Domain Finetune + SF                        & 37.9 & 20.3 \\
\cline {2-5}
                                      & \multirow{ 4}{*}{Medical} & Baseline  & 26.8 & 19.8 \\
                                      \cline{3-5} 
                                      & & Ensemble Across Domains             & 29.8 & 20.0 \\
                                      & & Mixed-Domain Finetune               & 29.9 & 18.9 \\
                                      & & In-Domain Finetune                  & 31.4 & 17.3 \\
                                      & & In-Domain Finetune + SF             & 32.2 & 17.8 \\
\hline
\end{tabular}
\caption{\label{ja-en-results}
Detailed Ja $\rightarrow$ En in-domain parallel results. SF stands for shallow fusion.
}
\end{table*}

\begin{table*}
\centering
\begin{tabular}{|c|c|l|c|c|}
\hline
\textbf{Languages}  & \textbf{Domain} & \textbf{Model Description} &  \textbf{In-Domain} & \textbf{Original Domain} \\
\hline
\multirow{10}{*}{Zh $\rightarrow$ En} & \multirow{ 4}{*}{Consumer Electronic} & Baseline & 34.5 & 24.5 \\
                                      \cline{3-5} 
                                      & & Ensemble Across Domains                        & 39.8 & 22.1 \\
                                      & & Mixed-Domain Finetune                          & 41.0 & 20.3 \\
                                      & & In-Domain Finetune                             & 42.1 & 14.2 \\
                                      & & In-Domain Finetune + SF                        & 42.2 & 14.1 \\
\cline {2-5}
                                      & \multirow{ 4}{*}{Medical} & Baseline  & 29.9 & 24.5 \\
                                      \cline{3-5} 
                                      & & Ensemble Across Domains             & 41.0 & 22.1 \\
                                      & & Mixed-Domain Finetune               & 44.8 & 20.7 \\
                                      & & In-Domain Finetune                  & 44.7 & 14.4 \\
                                      & & In-Domain Finetune + SF             & 45.0 & 19.5 \\
\hline
\end{tabular}
\caption{\label{zh-en-results}
Detailed Zh $\rightarrow$ En in-domain parallel results. SF stands for shallow fusion.
}
\end{table*}

\begin{table*}
\centering
\begin{tabular}{|c|c|l|c|c|}
\hline
\textbf{Languages}  & \textbf{Domain} & \textbf{Model Description} &  \textbf{In-Domain} & \textbf{Original Domain} \\
\hline
\multirow{10}{*}{Es $\rightarrow$ En} & \multirow{ 4}{*}{Consumer Electronic} & Baseline & 46.1 & 39.9 \\
                                      \cline{3-5}
                                      & & Ensemble Across Domains                        & 51.8 & 39.5 \\
                                      & & Mixed-Domain Finetune                          & 54.6 & 37.6 \\
                                      & & In-Domain Finetune                             & 56.4 & 33.7 \\
                                      & & In-Domain Finetune + SF                        & 56.6 & 33.7 \\
\cline {2-5}
                                      & \multirow{ 4}{*}{Medical} & Baseline  & 50.1 & 39.9 \\
                                      \cline{3-5}
                                      & & Ensemble Across Domains             & 54.1 & 39.5 \\
                                      & & Mixed-Domain Finetune               & 55.2 & 37.7 \\
                                      & & In-Domain Finetune                  & 55.3 & 36.5 \\
                                      & & In-Domain Finetune + SF             & 55.2 & 36.1 \\
\hline
\end{tabular}
\caption{\label{es-en-results}
Detailed Es $\rightarrow$ En in-domain parallel results. SF stands for shallow fusion.
}
\end{table*}

\begin{table*}
\centering
\begin{tabular}{|c|c|l|c|c|}
\hline
\textbf{Languages}  & \textbf{Domain} & \textbf{Model Description} &  \textbf{In-Domain} & \textbf{Original Domain} \\
\hline
\multirow{15}{*}{Ru $\rightarrow$ En} & \multirow{ 5}{*}{Consumer Electronic} & Baseline & 25.6 & 36.2 \\
                                      \cline{3-5}
                                      & & Ensemble Across Domains                        & 29.5 & 35.9 \\
                                      & & Mixed-Domain Finetune                          & 35.5 & 31.9 \\
                                      & & Mixed-Domain Finetune + SF                     & 35.8 & 32.2 \\
                                      & & In-Domain Finetune                             & 35.9 & 23.6 \\
                                      & & In-Domain Finetune + SF                        & 36.1 & 23.2 \\
                                      
\cline {2-5}
                                      & \multirow{ 5}{*}{Medical} & Baseline  & 27.7 & 36.2 \\
                                      \cline{3-5}
                                      & & Ensemble Across Domains             & 31.9 & 35.9 \\
                                      & & Mixed-Domain Finetune               & 39.2 & 32.3 \\
                                      & & Mixed-Domain Finetune + SF          & 39.4 & 32.5 \\
                                      & & In-Domain Finetune                  & 38.7 & 31.6 \\
                                      & & In-Domain Finetune + SF             & 39.2 & 31.8 \\
\cline {2-5}
                                      & \multirow{ 5}{*}{Biomedical} & Baseline  & 38.5 & 36.2 \\
                                      \cline{3-5}
                                      & & Ensemble Across Domains                & 39.0 & 35.9 \\
                                      & & Mixed-Domain Finetune                  & 41.3 & 37.0 \\
                                      & & Mixed-Domain Finetune + SF             & 41.6 & 37.1 \\
                                      & & In-Domain Finetune                     & 42.0 & 32.8 \\
                                      & & In-Domain Finetune + SF                & 41.7 & 32.4 \\
\hline
\end{tabular}
\caption{\label{Ru-Parallel-Data-Only}
Detailed Ru $\rightarrow$ En in-domain parallel results. SF stands for shallow fusion.
}
\end{table*}

\begin{table*}
\centering
\begin{tabular}{|c|c|l|c|c|}
\hline
\textbf{Languages}  & \textbf{Domain} & \textbf{Model Description} &  \textbf{In-Domain} & \textbf{Original Domain} \\
\hline
\multirow{12}{*}{Ru $\rightarrow$ En} & \multirow{ 5}{*}{Consumer Electronic} & Baseline   & 25.6 & 36.2 \\
                                      \cline{3-5}
                                      & & In-Domain + Baseline BT                          & 32.4 & 33.3 \\
                                      & & In-Domain + Finetuned BT                         & 34.4 & 25.8 \\
                                      & & In-Domain + Tagged Finetuned BT                  & 34.2 & 21.8 \\
                                      & & In-Domain + Tagged Finetuned BT + SF             & 34.8 & 22.1 \\
\cline {2-5}
                                      & \multirow{ 5}{*}{Medical} & Baseline               & 27.7 & 36.2 \\
                                      \cline{3-5}
                                      & & In-Domain + Baseline BT                          & 36.8 & 26.2 \\
                                      & & In-Domain + Finetuned BT                         & 37.3 & 27.1 \\
                                      & & In-Domain + Tagged Finetuned BT                  & 37.9 & 20.2 \\
                                      & & In-Domain + Tagged Finetuned BT + SF             & 38.2 & 20.0 \\
\cline {2-5}
                                      & \multirow{ 5}{*}{Biomedical} & Baseline            & 38.5 & 36.2 \\
                                      \cline{3-5}
                                      & & In-Domain + Baseline BT                          & 41.1 & 33.8 \\
                                      & & In-Domain + Finetuned BT                         & 40.9 & 34.6 \\
                                      & & In-Domain + Tagged Finetuned BT                  & 40.2 & 34.6 \\
                                      & & In-Domain + Tagged Finetuned BT + SF             & 41.0 & 34.8 \\
\hline
\end{tabular}
\caption{\label{ru-en-results}
Detailed Ru $\rightarrow$ En in-domain parallel + target monolingual results. BT stands for backtranslation and SF stands for shallow fusion.
}
\end{table*}

\begin{table*}
\centering
\begin{tabular}{|c|c|l|c|c|}
\hline
\textbf{Languages}  & \textbf{Domain} & \textbf{Model Description} &  \textbf{In-Domain} & \textbf{Original Domain} \\
\hline
\multirow{16}{*}{Ja $\rightarrow$ En} & \multirow{ 8}{*}{Consumer Electronic} & Baseline & 36.1 & 19.8 \\
                                      \cline{3-5}
                                      & & Baseline + SF                                  & 37.9 & 20.3 \\
                                      & & BT Top 200k                                    & 34.7 & 18.6 \\
                                      & & BT Top 50k                                     & 34.8 & 17.0 \\
                                      & & BT Top 50k + SF                                & 35.4 & 16.7 \\
                                      & & BT Top CE Total                                & 34.2 & 17.4 \\
                                      & & BT CE Target                                   & 36.3 & 17.6 \\
                                      & & BT CE Target + SF                              & 37.6 & 18.1 \\
\cline {2-5}
                                      & \multirow{ 8}{*}{Medical} & Baseline & 26.8 & 19.8 \\
                                      \cline{3-5}
                                      & & Baseline + SF                      & 29.2 & 20.5 \\
                                      & & BT Top 200k                        & 27.3 & 16.2 \\
                                      & & BT Top 50k                         & 27.3 & 16.5 \\
                                      & & BT Top 50k + SF                    & 29.3 & 18.0 \\
                                      & & BT Top Medical Total               & 27.5 & 15.5 \\
                                      & & BT Medical Target                  & 29.3 & 16.6 \\
                                      & & BT Medical Target + SF             & 31.4 & 16.9 \\
\hline
\end{tabular}
\caption{\label{Ja-Target-Data-Only}
Detailed Ja $\rightarrow$ En in-domain target monolingual results. BT stands for backtranslation and SF stands for shallow fusion.
}
\end{table*}

\begin{table*}
\centering
\begin{tabular}{|c|c|l|c|c|}
\hline
\textbf{Languages}  & \textbf{Domain} & \textbf{Model Description} &  \textbf{In-Domain} & \textbf{Original Domain} \\
\hline
\multirow{16}{*}{Zh $\rightarrow$ En} & \multirow{ 8}{*}{Consumer Electronic} & Baseline & 34.5 & 24.5 \\
                                      \cline{3-5}
                                      & & Baseline + SF                                  & 34.5 & 23.8 \\
                                      & & BT Top 200k                                    & 35.5 & 25.2 \\
                                      & & BT Top 50k                                     & 35.5 & 25.2 \\
                                      & & BT Top 50k + SF                                & 35.5 & 24.2 \\
                                      & & BT Top CE Total                                & 35.8 & 25.1 \\
                                      & & BT CE Target                                   & 38.2 & 26.2 \\
                                      & & BT CE Target + SF                              & 38.4 & 24.7 \\
\cline {2-5}
                                      & \multirow{ 8}{*}{Medical} & Baseline & 29.9 & 24.5 \\
                                      \cline{3-5}
                                      & & Baseline + SF                      & 29.7 & 20.2 \\
                                      & & BT Top 200k                        & 33.6 & 24.8 \\
                                      & & BT Top 50k                         & 35.6 & 17.2 \\
                                      & & BT Top 50k + SF                    & 36.2 & 15.5 \\
                                      & & BT Top Medical Total               & 34.6 & 20.1 \\
                                      & & BT Medical Target                  & 39.2 & 20.1 \\
                                      & & BT Medical Target + SF             & 42.0 & 19.5 \\
\hline
\end{tabular}
\caption{\label{Zh-Target-Data-Only}
Detailed Zh $\rightarrow$ En in-domain target monolingual results. BT stands for backtranslation and SF stands for shallow fusion.
}
\end{table*}

\begin{table*}
\centering
\begin{tabular}{|c|c|l|c|c|}
\hline
\textbf{Languages}  & \textbf{Domain} & \textbf{Model Description} &  \textbf{In-Domain} & \textbf{Original Domain} \\
\hline
\multirow{16}{*}{Es $\rightarrow$ En} & \multirow{ 8}{*}{Consumer Electronic} & Baseline & 46.1 & 39.9 \\
                                      \cline{3-5}
                                      & & Baseline + SF                                  & 46.7 & 40.0 \\
                                      & & BT Top 200k                                    & 46.8 & 38.6 \\
                                      & & BT Top 50k                                     & 47.2 & 35.8 \\
                                      & & BT Top 50k + SF                                & 48.1 & 36.3 \\
                                      & & BT Top CE Total                                & 48.3 & 39.8 \\
                                      & & BT CE Target                                   & 53.2 & 35.8 \\
                                      & & BT CE Target + SF                              & 53.3 & 35.9 \\
\cline {2-5}
                                      & \multirow{ 8}{*}{Medical} & Baseline & 50.1 & 39.9 \\
                                      \cline{3-5}
                                      & & Baseline + SF                      & 50.8 & 40.1 \\
                                      & & BT Top 200k                        & 49.3 & 35.5 \\
                                      & & BT Top 50k                         & 50.0 & 37.2 \\
                                      & & BT Top 50k + SF                    & 50.9 & 37.9 \\
                                      & & BT Top Medical Total               & 50.2 & 39.9 \\
                                      & & BT Medical Target                  & 52.5 & 34.8 \\
                                      & & BT Medical Target + SF             & 52.7 & 34.8 \\
\hline
\end{tabular}
\caption{\label{Es-Target-Data-Only}
Detailed Es $\rightarrow$ En in-domain target monolingual results. BT stands for backtranslation and SF stands for shallow fusion. 
}
\end{table*}

\begin{table*}
\centering
\begin{tabular}{|c|c|l|c|c|}
\hline
\textbf{Languages}  & \textbf{Domain} & \textbf{Model Description} &  \textbf{In-Domain} & \textbf{Original Domain} \\
\hline
\multirow{19}{*}{Ru $\rightarrow$ En} & \multirow{ 8}{*}{Consumer Electronic} & Baseline & 25.6 & 36.2 \\
                                      \cline{3-5}
                                      & & Baseline + SF                                  & 26.5 & 36.9 \\
                                      & & BT Top 200k                                    & 27.4 & 36.2 \\
                                      & & BT Top 50k                                     & 28.0 & 35.4 \\
                                      & & BT Top 50k + SF                                & 28.4 & 35.5 \\
                                      & & BT Top CE Total                                & 27.2 & 36.6 \\
                                      & & BT CE Target                                   & 30.5 & 32.2 \\
                                      & & BT CE Target + SF                              & 31.0 & 32.4 \\
\cline {2-5}
                                      & \multirow{ 8}{*}{Medical} & Baseline & 27.7 & 36.2 \\
                                      \cline{3-5}
                                      & & Baseline + SF                      & 28.4 & 37.1 \\
                                      & & BT Top 200k                        & 28.6 & 32.0 \\
                                      & & BT Top 50k                         & 28.5 & 34.3 \\
                                      & & BT Top 50k + SF                    & 29.8 & 34.5 \\
                                      & & BT Top Medical Total               & 28.4 & 36.6 \\
                                      & & BT Medical Target                  & 32.9 & 35.4 \\
                                      & & BT Medical Target + SF             & 33.4 & 35.6 \\
\cline {2-5}
                                      & \multirow{ 2}{*}{Biomedical} & Baseline & 38.5 & 36.2 \\
                                      \cline{3-5}
                                      & & Baseline + SF                         & 39.0 & 36.6 \\
\hline
\end{tabular}
\caption{\label{Ru-Target-Data-Only}
Detailed Ru $\rightarrow$ En in-domain target monolingual results. BT stands for backtranslation and SF stands for shallow fusion. 
}
\end{table*}


\end{document}